\documentclass{article}
\usepackage{graphicx, amsmath, amssymb}

\usepackage{subfig}
\usepackage[table,xcdraw]{xcolor}
\usepackage{booktabs} 
\usepackage{microtype}
\usepackage{multirow}
\usepackage{hologo} 
\title{\textbf{Loss Estimators Improve Model Generalization}}
\author{Vivek Narayanaswamy$^{\dagger}$, Jayaraman J. Thiagarajan$^{\ddagger}$\thanks{This work was performed under the auspices of the U.S. Department of Energy by Lawrence Livermore National Laboratory under Contract DE-AC52-07NA27344.}, Deepta Rajan$^+$, \\ Andreas Spanias$^{\dagger}$ \\
\textit{$^{\dagger}$Arizona State University, $^{\ddagger}$Lawrence Livermore National Laboratory}, \\ $^+$\textit{IBM Research Almaden}
}
\date{}
\begin{document}
\maketitle
\begin{abstract}
With increased interest in adopting AI methods for clinical diagnosis, a vital step towards safe deployment of such tools is to ensure that the models not only produce accurate predictions but also do not generalize to data regimes where the training data provide no meaningful evidence. Existing approaches for ensuring the distribution of model predictions to be similar to that of the true distribution rely on explicit uncertainty estimators that are inherently hard to calibrate. In this paper, we propose to train a \textit{loss estimator} alongside the predictive model, using a contrastive training objective, to directly estimate the prediction uncertainties. Interestingly, we find that, in addition to producing well-calibrated uncertainties, this approach improves the generalization behavior of the predictor. Using a dermatology use-case, we show the impact of loss estimators on model generalization, in terms of both its fidelity on in-distribution data and its ability to detect out of distribution samples or new classes unseen during training.


\end{abstract}

\section{Introduction}
\label{sec:intro}
Deep learning methods are routinely used in state-of-the-art AI models for clinical diagnosis with imaging data. In particular, over the last few years, we have witnessed critical advances to the use of AI in radiology~\cite{hosny2018artificial} and dermatology~\cite{young2020artificial}. A key step towards promoting the adoption of these tools in practice is to ensure that the models behave predictably, and do not provide unintended generalization to regimes where the training data provide no meaningful evidence. In this paper, our focus is on studying the generalization behavior of deep networks using a dermatology use-case, both in terms of making accurate predictions for samples from the training data distribution as well as detecting new classes or out of distribution (OOD) samples~\cite{cao2020benchmark}. 

The problem of accurately identifying lesion types from dermatology images is known to be challenging and hence has been featured in a number of competitions~\cite{isic}. It is important to note that a model that produces high accuracy on the original data distribution is not always guaranteed to be effective at detecting OOD samples. On the other hand, models that are overly sensitive to even mild distribution shifts can provide inferior in-distribution performance. This clearly emphasizes the need for learning strategies that can effectively regulate the predictions, such that the model is well-calibrated to detect OOD samples (and can defer from making predictions), while not trading off performance.  

A popular approach for ensuring that the distribution of model prediction probabilities matches the true distribution from the observed data is to perform post-hoc calibration using strategies such as Platt scaling~\cite{guo2017calibration} and isotonic regression~\cite{kuleshov2018accurate} based on a validation dataset. On the other hand, approaches that temper model confidences based on explicit uncertainty estimators have also been proposed~\cite{seo2019learning,thiagarajan2020building}. While the effectiveness of the former approach relies heavily on the choice of the validation dataset, epistemic uncertainty estimators used in the latter class of approaches are found to be poorly calibrated in practice~\cite{kuleshov2018accurate}. In general, epistemic uncertainty refers to the lack of knowledge and could be eliminated with sufficient data -- such an optimal learner is referred to as the Bayes optimal predictor. The irreducible error associated with the Bayes optimal predictor is the aleatoric uncertainty. 

\begin{figure}[t]
\includegraphics[width=0.99\textwidth]{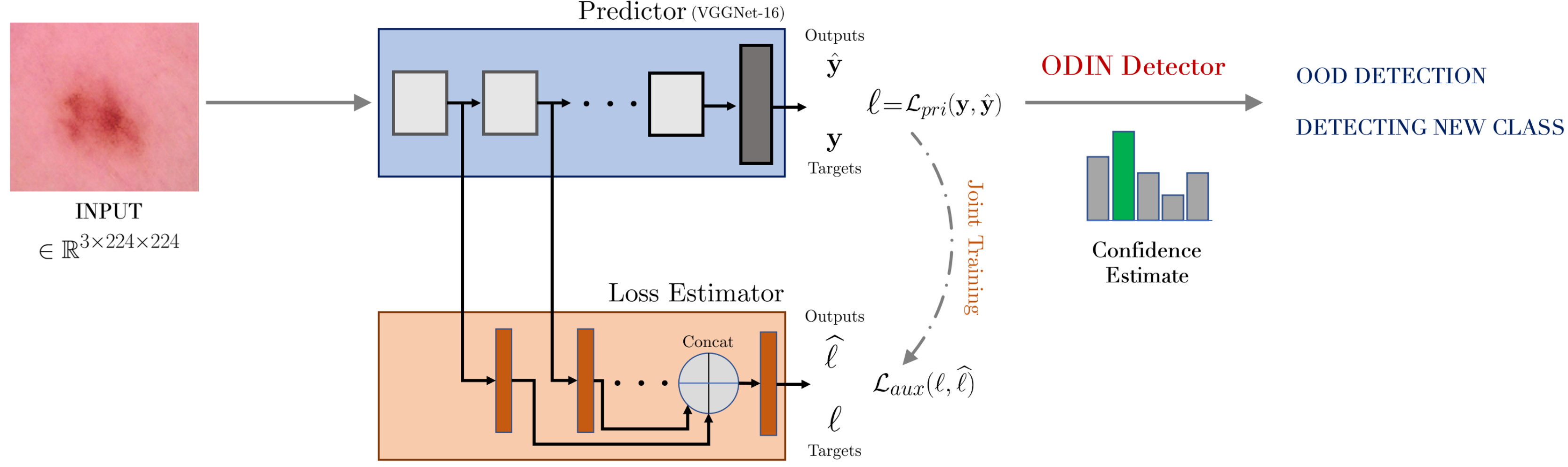}
\caption{An overview of our proposed approach. We propose to jointly train a loss estimator alongside the classifier in order to obtain the prediction uncertainties. This joint training process in turn regularizes the classifier model training and helps produce well-calibrated predictions. Finally, we use the ODIN detector~\cite{liang2018enhancing} to perform OOD detection with the trained predictor.} \label{fig:blockdiag}
\end{figure}

In this paper, we propose to employ a \textit{loss estimator} to directly predict the generalization error (sum of both epistemic and aleatoric uncertainties) and thereby produce inherently calibrated uncertainty estimators. While such a loss estimator can be learned for any pre-trained predictor~\cite{jain2021deup}, we make an interesting finding that jointly training the loss estimator alongside a predictor improves the generalization behavior of the predictor, in addition to producing calibrated uncertainties. Our contributions can be summarized as follows: (i) We perform uncertainty estimation based on a loss estimator, which is trained alongside the predictor using a contrastive training objective; (ii) Using the ISIC 2019 challenge dataset~\cite{codella2018skin}, we demonstrate improved model generalization to in-distribution data using standard classification metrics; (iii) Using a state-of-the-art OOD detector~\cite{liang2018enhancing}, we show that our predictor achieves significant improvements in detecting out of distribution samples; (iv) Finally, we find that predictors trained by coupling a loss estimator can more effectively detect new classes not seen during training.

\section{Problem Setup}
\label{sec:setup}
Our goal is to build a functional mapping $\mathcal{F}$ between an input image $\mathrm{x}$ and the output $\mathrm{y} \in \mathcal{Y}=\{1,2,\cdots,K\}$, which corresponds to a $K-$way classification task. We consider the standard supervised learning setup, where the expected discrepancy between $\mathrm{y}$ and $\mathcal{F}(\mathrm{x})$, measured using a loss function $\mathcal{L}(\mathrm{y}, \mathcal{F}(\mathrm{x}))$, is minimized over the joint distribution $p(\mathrm{x}, \mathrm{y})$. The predictive modeling process amounts to estimating the tuple $(\hat{\mathrm{y}}, \hat{p})$, where $\hat{\mathrm{y}}$ is the predicted label and $\hat{p}$ is the likelihood. As discussed in Section \ref{sec:intro}, our goal is to improve the generalization of $\mathcal{F}$ by making accurate predictions for in-distribution data and reliably detecting OOD data. Such generalization characteristics are critical for reliably deploying clinical predictive models.

In this study, we consider the problem of dermoscopic lesion classification and utilized the  ISIC 2019 lesion diagnosis challenge dataset containing 25,331 images acquired and consolidated from benchmark datasets provided by ~\cite{tschandl2018ham10000,codella2018skin,combalia2019bcn20000}. Each image in this dataset belongs to one of the following $8$ categories: Melanoma (MEL), Melanocytic nevus (NV), Basal cell carcinoma (BCC), Actinic keratosis (AK), Benign keratosis (BKL), Dermatofibroma (DF), Vascular lesion (VASC) and Squamous cell carcinoma (SCC). Note that, similar to several existing medical image classification benchmarks, this dataset is also characterized by class imbalances and large intra-class variances.



\section{Proposed Approach}
\label{sec:approach}
\begin{figure}[t]
\includegraphics[width=0.99\textwidth]{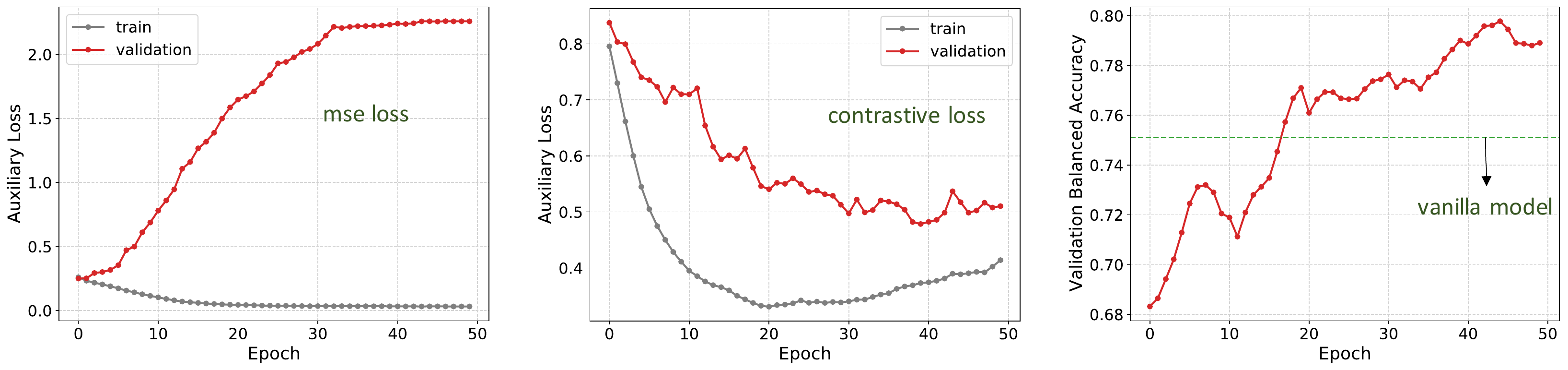}
\caption{Learning behavior of the proposed joint training process. Using a contrastive objective for the auxiliary loss makes the loss estimator generalize to the validation data (middle) and regularizes the predictor to improve its accuracy (right). In contrast, implementing the auxiliary loss as the standard mse objective provides very poor generalization to unseen data (left).} \label{fig:training}
\end{figure}

Figure \ref{fig:blockdiag} illustrates an overview of the proposed approach for improving the generalization of medical image classifiers. Our key idea is to include a prediction uncertainty estimator into the training process. More specifically, we build a loss estimator that directly predicts the generalization error of the predictor. Note that, recently in~\cite{jain2021deup}, a similar direct uncertainty estimation strategy was proposed, but with a key difference that the estimator was trained separately for a pre-trained predictor model. In contrast, our approach jointly trains both models thereby leveraging this calibration process as a regularization for improving generalization.    
Formally, we treat the lesion type prediction as the primary task and generalization error estimation as the auxiliary task. Assuming that the model $\mathcal{F}$ is trained to optimize the cross entropy loss $\mathcal{L}_{pri} = \mathcal{L}_{CE}(\mathcal{F}(\mathrm{x}), \mathrm{y})$, we construct the loss estimator $\mathcal{G}$ to estimate the loss $\mathrm{\ell} = \mathcal{L}_{pri}$. The architecture for $\mathcal{G}$ uses a linear layer with ReLU activation to transform representations from different layers of $\mathcal{F}$ (e.g., outputs from different convolutional layers) and employs a final linear layer on the concatenated features to produce the loss estimates. Now, the key challenge is to choose an appropriate optimization objective for the auxiliary task, $\mathcal{L}_{aux}(\mathrm{\ell}, \hat{\mathrm{\ell}})$. For example, one can use the mean squared error (mse) objective. However, as showed in Figure \ref{fig:training}(left), loss estimators trained with the mse objective do not generalize to new data, since the loss estimator converges to produce an average loss value for all images. Hence, we employ a contrastive loss which aims to preserve the ordering of samples based on their corresponding losses from $\mathcal{F}$. Let $\mathrm{\ell}_i$ and $\mathrm{\ell}_j$ denote the losses of samples $\mathrm{x}_i$ and $\mathrm{x}_j$, while the corresponding estimates from $\mathcal{G}$ are $\hat{\mathrm{\ell}}_i$ and $\hat{\mathrm{\ell}}_j$ respectively. Mathematically, 
\begin{align}
    \mathcal{L}_{aux} = &\sum_{(i,j)}\max \bigg(0, -\mathbb{I}(\mathrm{\ell}_i,\mathrm{\ell}_j) . (\hat{\mathrm{\ell}}_i - \hat{\mathrm{\ell}}_j) + \gamma \bigg), \\
    &\nonumber \text{where } \mathbb{I}(\mathrm{\ell}_i,\mathrm{\ell}_j) = \begin{cases}
1, &\text{if $\mathrm{\ell}_i > \mathrm{\ell}_j$},\\
-1, &\text{otherwise}.
\end{cases}
    \label{eqn:laux1}
\end{align}Here $\gamma$ is the margin parameter. For example, when the sign of $\mathrm{\ell}_i - \mathrm{\ell}_j$ is positive, we assign a non-zero penalty if the estimates $\hat{\mathrm{\ell}}_j > \hat{\mathrm{\ell}}_i$, i.e., there is a disagreement in the ranking of samples.The overall objective for the joint optimization of the predictor and loss estimator is given by
\begin{equation}
    \mathcal{L}_{total} = \mathcal{L}_{pri} + \lambda \mathcal{L}_{aux}. 
\end{equation}The parameter $\lambda$ was set to $0.5$ in all our experiments. Interestingly, as showed in Figure \ref{fig:training}, using the contrastive objective produces generalizable loss estimators and also effectively regularizes the predictor model training (indicated by significant improvement in the validation accuracy over an unregularized model).

\section{Results and Findings}
\label{sec:results}
We performed empirical studies to evaluate the proposed approach in model generalization, OOD detection, and discovering new classes unseen during training.
\subsection{Generalization Performance}
In this experiment, we build predictor models with and without an additional \textit{loss estimator} module (Fig. ~\ref{fig:blockdiag}), and present comparisons on the ISIC 2019 dataset.

\begin{figure}[t]
\includegraphics[width=0.99\textwidth]{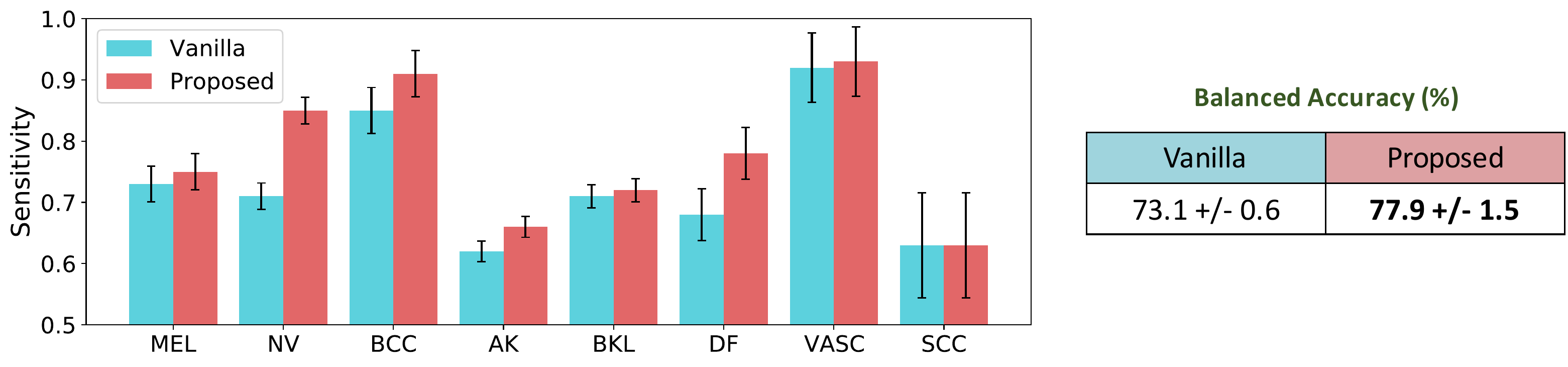}
\caption{Performance comparison of the vanilla and proposed models using sensitivity and balanced accuracy metrics (results averaged across $3$ independent trials) on the ISIC 2019 lesion dataset.} \label{fig:gen}
\end{figure}

\paragraph{\textbf{Setup}}.  We learned the predictor models using the ISIC 2019 dataset (Task1) -- we used $90-10$ stratified random splits for train/validation and report performance from $3$ independent trials. We adopted standard data pre-processing steps used for this data~\cite{gessert2020skin} -- all images were center cropped to $0.85$ times its original width, adjusted for color constancy and resized to $224 \times 224 \times 3$. Further, we performed data augmentation (horizontal/vertical flips, jitter, rotation and translation). For all experiments, we adopted a  VGG-16 network (pre-trained on imagenet) and fine-tuned it for $50$ epochs. The loss estimator utilized the activations of the intermediate convolutional layers $4$, $7$, $10$ and $13$ from the predictor followed by affine transformations using linear layers of $128$ units each. These individual representations are then concatenated and transformed using a final linear layer to obtain the loss estimate for that sample. For model training, we used the following hyperparameters: learning rate of $1e-4$ reduced by a factor of $0.5$ every $10$ epochs, Adam optimizer with $0.9$ momentum and $5e-4$ weight decay, dropout of $0.4$, and batch size of $64$. In addition, to handle the class imbalances and perform well on undersampled classes (e.g. AK, DF, BCC), we used weighted random sampling and a weighted cross-entropy loss. Following common practice, we compared the models using the class-specific sensitivity and balanced accuracy scores.
\paragraph{\textbf{Findings}}. As showed in Figure \ref{fig:gen}, our approach convincingly outperforms the vanilla model in classifying in-distribution data. Jointly training the loss estimator alongside the predictor (i) allows training with higher learning rates (faster convergence), (ii) achieves higher performance on minority classes (BCC, AK, DF), while also improving performance for densely sampled classes (NV, MEL), and (iii) produces $\sim 5$\% improvement over the vanilla model in terms of balanced accuracy score ($77.9$\% as opposed to $73.1$\% of the vanilla model).

\subsection{OOD Detection}
In this section, we setup an experiment to study and evaluate the effectiveness of the loss estimator-regularized predictor in detecting out-of-distribution images. We begin by describing the training setup, the OOD detector implemented following~\cite{liang2018enhancing}, and the metrics utilized to for evaluation.

\paragraph{\textbf{Setup.}} In this study, we utilized 5 out of the 8 lesion image categories from the ISIC 2019 training dataset, namely MEL, NV, BCC, AK and DF, to define the in-distribution data. This selection amounts to a total of $21,826$ samples. We performed a $90 - 10$  stratified random split of images from these 5 classes to train and validate the models. We used the same architectures as the previous experiment to implement the predictor and loss estimator models. Note, we used the the ADAM optimizer with a learning rate of $1e-4$ and trained the models for $50$ epochs. For comparison, we also trained a vanilla model under the same experiment settings.

\paragraph{\textbf{ODIN Detector.}}In order to evaluate the model reliability in rejecting samples (defer to make predictions) when presented images from distinctly different distributions, we adopt ODIN (Out of DIstribution detector in Neural networks)~\cite{liang2018enhancing}, a popular approach for detecting OOD data based upon the calibrated confidence (softmax) scores. In particular, ODIN employs temperature scaling to the softmax probabilities and applies controlled input image perturbations to enlarge the confidence score gap between the in-distribution and out-of-distribution data. For an input sample $\mathbf{x}$, ODIN first computes the confidence score from temperature~($T$) scaled softmax probabilities:
\begin{equation}
    S(\mathbf{x}, T) = \max_{c} \frac{\exp({\mathcal{F}_{c}(\mathbf{x}})/T)}{\sum_{k=1}^{K}\exp({\mathcal{F}_{k}(\mathbf{x}})/T)}
\end{equation}
where $\mathcal{F}_{c}(\mathbf{x})$ is the predictive model output for the $c^{th}$ class in a $K$-way classification problem. In addition to temperature scaling, ODIN also systematically perturbs the input by a factor of $\eta$ in the direction of the gradient of the loss w.r.t $\mathbf{x}$, in order to improve the softmax probabilities. Formally,
\begin{equation}
    \hat{\mathbf{x}} = \mathbf{x} - \eta \text{ sign}(-\nabla_{\mathbf{x}}\text{log}(S(\mathbf{x}, T)) 
\end{equation}Finally, the confidence score from the perturbed image is used to determine if $\mathrm{x}$ is an inlier sample --  based on a user-specified threshold $\gamma$.  

\begin{figure*}[t]
    \centering
    \subfloat[S][B-Box]{\includegraphics[width=0.16\textwidth]{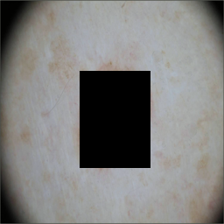}}
    \hfill
    \subfloat[S][B-Box-70]{\includegraphics[width=0.16\textwidth]{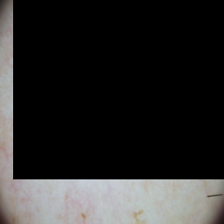}} 
    \hfill
    \subfloat[S][Imagenet]{\includegraphics[width=0.16\textwidth]{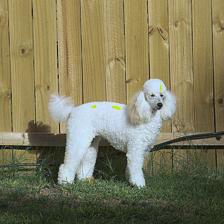}}
    \hfill
    \subfloat[S][NCT]{\includegraphics[width=0.16\textwidth]{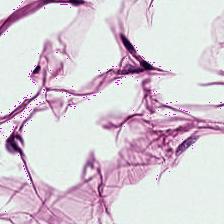}}
    \hfill
    \subfloat[S][Clin Skin]{\includegraphics[width=0.16\textwidth]{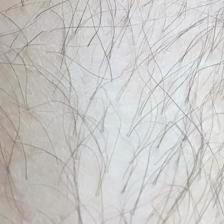}}
    \hfill
    \subfloat[S][Derm Skin]{\includegraphics[width=0.16\textwidth]{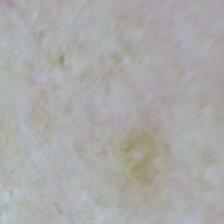}}
    \caption{Examples from the OOD Datasets used for our experiments.}
    \label{fig:ood_examples}
\end{figure*}
\begin{table}[t]
	\centering
	\renewcommand*{\arraystretch}{1.5}
	\caption{Evaluating the OOD detection performance of the ODIN detector using both the vanilla and proposed methods. The in-distribution data includes validation set samples belonging to the $5$ classes. The hyperparameters for ODIN are fine tuned on the NCT dataset and evaluated on the remaining datasets.}
	\resizebox{\textwidth}{!}{%
		\label{ood_perf}
		\begin{tabular}{|c|c|c|c|c|c|}
			\hline
			\multirow{3}{*}{\textbf{OOD Dataset}} & 
			\multicolumn{5}{c|}{\textbf{Metrics in \%}}
			\\ \cline{2-6}
			
			\multirow{1}{*}{} & \textbf{FPR@TPR95} & \textbf{DTERR} & \textbf{AUROC}  & \textbf{AUPR-In} & \textbf{AUPR-Out} \\
			\cline{2-6}
			
			\multirow{1}{*}{} & \multicolumn{5}{c|}{\textbf{Vanilla/Proposed}}
			\\ \cline{1-6}
			
			\multirow{1}{*}{NCT} & $51.41$/$\mathbf{17.93}$ & $21.44$/$\mathbf{10.19}$ & $87.27$/$\mathbf{96.43}$ & $91.41$/$\mathbf{97.73}$ & $81.71$/$\mathbf{94.43}$ 
			\\ \cline{1-6}
			
			\multirow{1}{*}{B-Box} & $36.43$/$\mathbf{5.16}$ & $17.84$/$\mathbf{4.91}$ & $90.74$/$\mathbf{98.74}$ & $90.91$/$\mathbf{98.71}$ & $91.15$/$\mathbf{98.71}$ 
			\\   \cline{1-6}
			
			\multirow{1}{*}{B-Box-70} & $\mathbf{0.0}$/$\mathbf{0.0}$ & $\mathbf{0.1}$/$\mathbf{0.1}$ & $\mathbf{100.0}$/$\mathbf{100.0}$ & $\mathbf{99.97}$/$\mathbf{99.97}$ & $\mathbf{99.98}$/$\mathbf{99.98}$ 
			\\ \cline{1-6}
			
			\multirow{1}{*}{ImageNet} & $71.05$/$\mathbf{47.35}$ & $28.7$/$\mathbf{20.92}$ & $79.03$/$\mathbf{87.57}$ & $81.68$/$\mathbf{87.74}$ & $76.38$/$\mathbf{86.84}$ 
			\\ \cline{1-6}
			
			\multirow{1}{*}{Clin-Skin} & $65.56$/$\mathbf{35.13}$ & $25.0$/$\mathbf{14.11}$ & $82.58$/$\mathbf{92.15}$ & $93.37$/$\mathbf{96.71}$ & $61.07$/$\mathbf{79.09}$ 
			\\ \cline{1-6}
			
			\multirow{1}{*}{Derm-Skin} & $56.04$/$\mathbf{21.6}$ & $21.6$/$\mathbf{10.38}$  & $87.14$/$\mathbf{95.69}$ & $91.22$/$\mathbf{97.2}$ & $82.59$/$\mathbf{92.14}$
			\\ \cline{1-6}
			
			 \hline
		\end{tabular}
	}
\end{table}

\paragraph{\textbf{OOD Datasets.}}In order to demonstrate the behavior of the predictor under a wide variety of unknown data regimes, we consider the following benchmark datasets, following~\cite{pacheco2020out}. In particular, we consider (i) \textit{B-Box}: $2025$ skin lesion images corrupted by a black bounding box on the lesion region; (ii) \textit{B-Box-70}: $2454$ skin lesion images with black bounding boxes that mask $\sim$ 70\% of the lesion region; (iii) \textit{ImageNet}~\cite{imagenet_cvpr09}: - $3000$ images randomly chosen from the ImageNet database; (iv) \textit{NCT}: $1350$ histopathology images of human colorectal cancer acquired from ~\cite{kather2019predicting} (v) \textit{Clin-Skin}: $723$ clinical images of healthy skin; (vi) \textit{Derm-Skin}: $1565$ dermoscopy images of skin obtained by randomly cropping patches in the ISIC2019 dataset. Figure \ref{fig:ood_examples} shows sample images from each of these datasets. 

\paragraph{\textbf{Evaluation Metrics.}} We adopt these widely adopted metrics to quantify the OOD detection performance: (i) \textbf{False Positive Rate $@95$\% True Positive Rate}(FPR@TPR95): Probability that an OOD sample is misclassified as an in-distribution sample when the TPR is as high as $95$ \%; (ii) \textbf{Detection Error}(DTERR): Minimum probability of mis-detecting an inlier sample as an OOD sample over all possible thresholds; (iii) \textbf{AUROC}: Area Under the Receiver Operator Characteristic curve (TPR vs FPR) is a threshold independent metric which reflects the probability that a in-distribution image is assigned a higher confidence over the OOD sample; (iv) \textbf{AUPR-In} and \textbf{AUPR-Out}: Area under the Precision-Recall curve where the in distribution and the OOD samples are considered as positives respectively.  

\paragraph{\textbf{Findings.}} In Table \ref{ood_perf}, we report the OOD detection performance of the proposed approach against the baseline vanilla predictor over a variety of datasets characterized by apparent distribution changes. The hyperparameters $T$ and $\eta$ were fine tuned for both the models using the NCT dataset and evaluated on the remaining datasets. It must be noted that while using the ODIN detector, we utilized the validation split of the 5 class ISIC 2019 dataset as the inlier distribution and the other listed datasets as OOD to compute the metrics. It can be observed from Table \ref{ood_perf} that the proposed model, which is effectively regularized by the loss estimator, significantly outperforms the baseline, thus implying improved model reliability under data regime changes. While it is more challenging to detect OOD samples which exhibit semantic similarities with the in-distribution data~\cite{cao2020benchmark}, e.g., \textit{Clin-Skin} and \textit{Derm-Skin}, we find that by coupling the uncertainty estimator during training, we can still effectively detect these shifts with improved specificity. 

\begin{figure*}[t]
    \centering
   \includegraphics[width=0.99\textwidth]{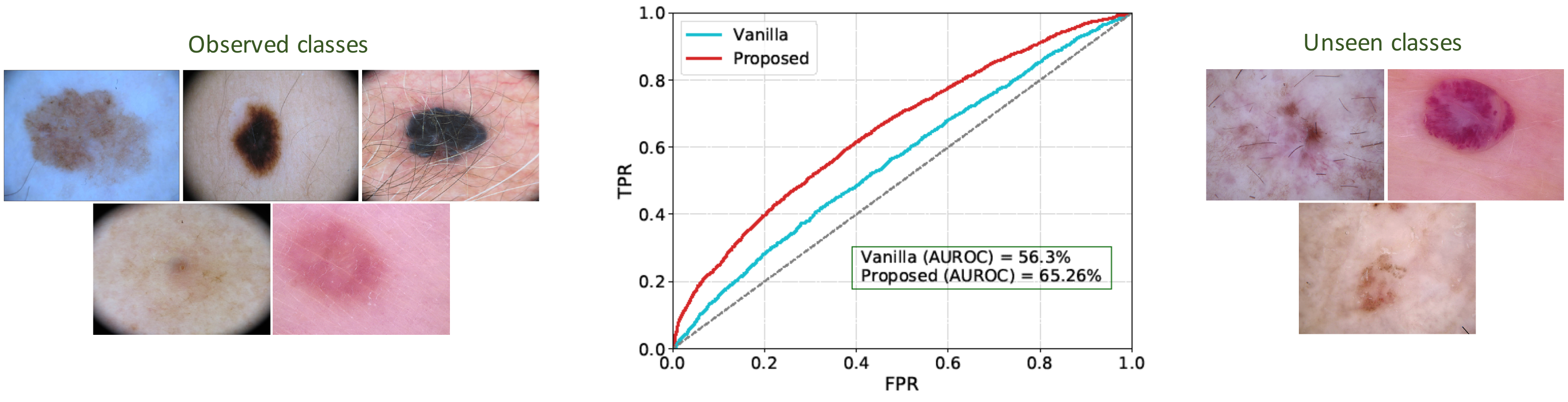}
    \caption{Performance of the proposed approach in detecting new classes unseen during training.}
    \label{fig:newclasses}
\end{figure*}

\subsection{Detecting New Classes}
In this experiment, we evaluate the ability of our proposed approach in detecting classes of data that are unseen during model training. Similar to the previous experiment, we utilized both the vanilla and the proposed models trained on the $5$ class ISIC 2019 dataset (Section 4.2). Subsequently, we introduced images from the $3$ remaining classes (BKL, VASC and SCC) as novel data at test time. We expect a reliable model to effectively detect these samples as OOD, thus enabling us defer from making an incorrect diagnosis. We find that the task of detecting new classes is significantly more challenging due to the less apparent semantic discrepancies between the observed and the unseen classes. From the results in Figure \ref{fig:newclasses}, we notice that the proposed approach still outperforms standard deep models (in terms of the AUROC metric -- Vanilla: $56.3$\%, Proposed : $65.26$\%) in detecting samples from unseen classes. This further emphasizes the value of loss estimators in controlling unintended generalization of clinical predictive models.

\section{Conclusions}
\label{sec:conclusion}
In this work, we showed that loss estimators are effective uncertainty predictors (generalization error) in deep neural networks, and more importantly, can regularize model training when trained jointly with the classifier. As an interesting byproduct, models obtained using this approach demonstrate strong generalization characteristics. Using a challenging dermoscopy image benchmark, we showed that loss estimators lead to more accurate classifiers, and more importantly, reliable OOD detection. Our study clearly emphasizes the role of calibration in controlling unintended generalization of clinical predictive models.

\bibliographystyle{splncs04}
\bibliography{refs}
\end{document}